\documentclass[conference]{IEEEtran}
\IEEEoverridecommandlockouts
\usepackage{cite}
\usepackage{amsmath,amssymb,amsfonts}
\usepackage{algorithmic}
\usepackage{graphicx}
\usepackage{textcomp}
\usepackage{xcolor}
\usepackage{subfig}
\usepackage{float}

\DeclareCaptionFont{mysize}{\fontsize{8}{9.6}\selectfont}
\def\BibTeX{{\rm B\kern-.05em{\sc i\kern-.025em b}\kern-.08em
    T\kern-.1667em\lower.7ex\hbox{E}\kern-.125emX}}
\begin{document}

\title{Convolutional Neural Network (CNN) vs Vision Transformer (ViT) for Digital Holography}

\author{\IEEEauthorblockN{1\textsuperscript{st} Stéphane Cuenat}
\IEEEauthorblockA{\textit{FEMTO-ST Institute, CNRS} \\
\textit{Univ. Bourgogne Franche-Comte (UBFC)}\\
Belfort, France \\
stephane.cuenat@univ-fcomte.fr \\
}
\and
\IEEEauthorblockN{2\textsuperscript{nd} Raphaël Couturier}
\IEEEauthorblockA{\textit{FEMTO-ST Institute, CNRS} \\
\textit{Univ. Bourgogne Franche-Comte (UBFC)}\\
Belfort, France \\
raphael.couturier@univ-fcomte.fr\\ 
}
}

\maketitle

\begin{abstract}

In Digital Holography (DH), it is crucial to extract the object distance from a hologram in order to reconstruct its amplitude and phase. This step is called auto-focusing and it is conventionally solved by first reconstructing a stack of images and then by sharpening each reconstructed image using a focus metric such as entropy or variance. The distance corresponding to the sharpest image is considered the focal position. This approach, while effective, is computationally demanding and time-consuming.  In this paper, the determination of the distance is performed by Deep Learning (DL). Two deep learning (DL) architectures are compared: Convolutional Neural Network (CNN) and Vision transformer (ViT).  ViT and CNN are used to cope with the problem of auto-focusing as a classification problem. Compared to a first attempt~\cite{b1} in which the distance between two consecutive classes was $100\mu m$, our proposal allows us to drastically reduce this distance to $1 \mu m$. Moreover, ViT reaches similar accuracy and is more robust than CNN.  
\end{abstract}

\begin{IEEEkeywords}
Digital holography, CNN, Efficientnet, Densenet, Vision Transformer, ViT, Transformer
\end{IEEEkeywords}

\section{Introduction}
Digital holography (DH) is an emerging field in imaging applications \cite{b11}. It is mostly exploited in 3D image processing, surface contour measurements, microscopy \cite{b2}, and even in microrobotics \cite{b3}. A major challenge in microrobotics and/or photonics is to be able to determine metrics in the context of complex imaging devices. Among those metrics, the 3D positioning of micro-objects is particularly interesting. With earlier hologram image reconstruction, the overall experimental setup needed to
be determined ahead of time including object’s depth position \cite{b4}, otherwise many diffraction calculations were needed to be applied for various depth settings. According to \cite{b5} and \cite{b6}, these techniques consume a lot of time as they necessitate many diffraction calculations and signal processing, and thus heavy computations were required in this scope. Nowadays, Deep Learning (DL) is reshaping the world of computer science and  is used in many application areas including DH. Particularly, DL helped in coping with time consumption and heavy computation concerns of the older techniques to determine depth position: instead of applying many diffraction calculations, training a deep neural network enabled to determine the depth predictions \cite{b7} and \cite{b8}. These predictions can be approached either as a classification problem \cite{b4} \cite{b9} \cite{b10} or as a regression problem \cite{b7} and \cite{b8}.

A first approach using a classification has been presented in~\cite{b1} showing that residual neural networks such as Densenet169 and Densenet201 can achieve good results and determine the auto-focusing distance with a precision of a scale of $100\mu$m. 

In this paper, the problem of determining the distance of a view captured by a holographic camera is addressed by
modeling it as a classification problem. The
outcomes proved that the built networks are capable of performing predictions with an accuracy level of 99\% at a micrometer scale, using 10 classes for
the classifications, where each class corresponds to a different distance from the object varying with a step of 1 micrometer (so this corresponds to a distance 100 times smaller than in~\cite{b1}).
The proposal is presented thereafter throughout the following sections. Section 2 presents some of the existing literature, more specifically on deep learning
and its benefits for addressing classification problems in digital holography. The context of this study is highlighted in the following section. Section 4 is dedicated to a detailed presentation of the proposed deep neural networks and the experimental results. Finally, a conclusion aims to summarize the findings and the potential future works.

\section{Related work}
First, DL is approached globally before digging deeper into its use in DH, then a presentation of what a transformer model means and how this architecture has been applied to classification problems is given. A Convolutional Neural Network (CNN) is a well-known DL architecture that is widely used for
analyzing and classifying images by extracting and learning features directly
from them \cite{b12}. Many CNN models are available, each having its own particularity
and advantages. Densenet \cite{b13} (a densely connected network) is one of them, 
it is known for its significant results when compared to other models. It
even necessitates fewer parameters for training \cite{b14}. According to the authors of
this CNN model \cite{b13}, Densenet networks do not encounter optimization issues,
even when scaling to hundreds of layers. Their main motivation to create this
model was to cope with the vanishing of the input information that occurs at the
output of the network after it has passed through many layers, as well as the
vanishing of the gradient in the opposite direction.
In the following,  some works in the literature that tackle DH
using DL are discussed. Many studies in DH focus on reconstructing the target
object’s amplitude and phase once its distance is known; a process referred
to as “autofocusing”. In particular, identifying the object distance is crucial for
the object reconstruction \cite{b4}. In the literature, autofocusing has been explored
by exploiting advances in DL (also called learning-based approaches). To be
more specific, two main approaches are investigated to tackle the problem: using
classification \cite{b4} \cite{b9} \cite{b10} and using regression \cite{b7} and \cite{b8}.

The focus is first put on the classification approach and its adoption in the literature.
In this approach, \cite{b9} and \cite{b10} were the first to work on predicting depth in DH
in discrete values. In \cite{b9}, the authors demonstrated that autofocusing in DH can
be achieved using DL. They adopted a CNN that is built on top of the AlexNet
architecture. They started off with a number of holograms that they prepared
first  by eliminating the “zero-order” and the “twin-term”. They used
21 classes for the labeling and they applied data augmentation on the prepared
holograms to increase the size of the dataset. Moreover, the network was trained
with manually defined depth values. 90\% of the data was used to train the
network and the remaining 10\% for the validation. Their learning rate was set
up to the value of 0.01. As they concluded, their network did not scale properly.
Moreover, they stated that their network “generalized well” at a millimeter scale
without mentioning the obtained accuracy level to prove it.
The authors of [13] used a CNN to predict the object distance by training
their network with hologram-specific labels that correspond to the actual distances. They used a uniform technical setup to capture 1,000 holograms, ending
up with 5 distance labels. Their experimental results proved that the network is
able to predict the distance without any knowledge of the technical setup and
with less time consumption than the traditional methods. This proves that they
coped with the issues that were encountered in the work \cite{b9}. 
However, although they
provided estimates at a millimeter scale (axial range of =3 mm, which is appropriate for imaging systems), they only worked on 5 classes, which may not be
enough to generalize their findings.

As shown in \cite{b1}, a first step has been reached using either Densenet169 or Densenet201. To train the model, 10,101 images were used with a split rule of 70\% / 30\%. All the images (hologram) used were simulated holograms. 

Vision Transformer architecture (ViT) has been introduced with the advent of the transformer architecture, mainly used in the field of language translation (NLP). ViT follows the core concepts of a transformer network which is based on the concept of self-attention. As pointed in \cite{b16} "attention is all we need". Such a network is based on an encoder which is built using self attention map mechanism. Contrary to CNN, ViT does not imply any convolution layer. The input image is split in patches and fed into ViT network. ViT architecture achieved state-of-the-art performance on Imagenet.

In this paper, 2 different architecture (CNN and ViT) are used to tackle the problem of depth
prediction in the DH context. Each architecture classifies in 10 classes and at a micrometer
scale. It uses a dataset of 3,600 images that are augmented and split using the 80-20\% rule, without using the training images in the testing. The model is pre-trained using Imagenet. Experimental results presented thereafter prove that the network is able
to predict depth with 99\% of accuracy.

\section{Proposed solution}
\begin{figure}[b]
    \centering
    \includegraphics[width=0.8\columnwidth]{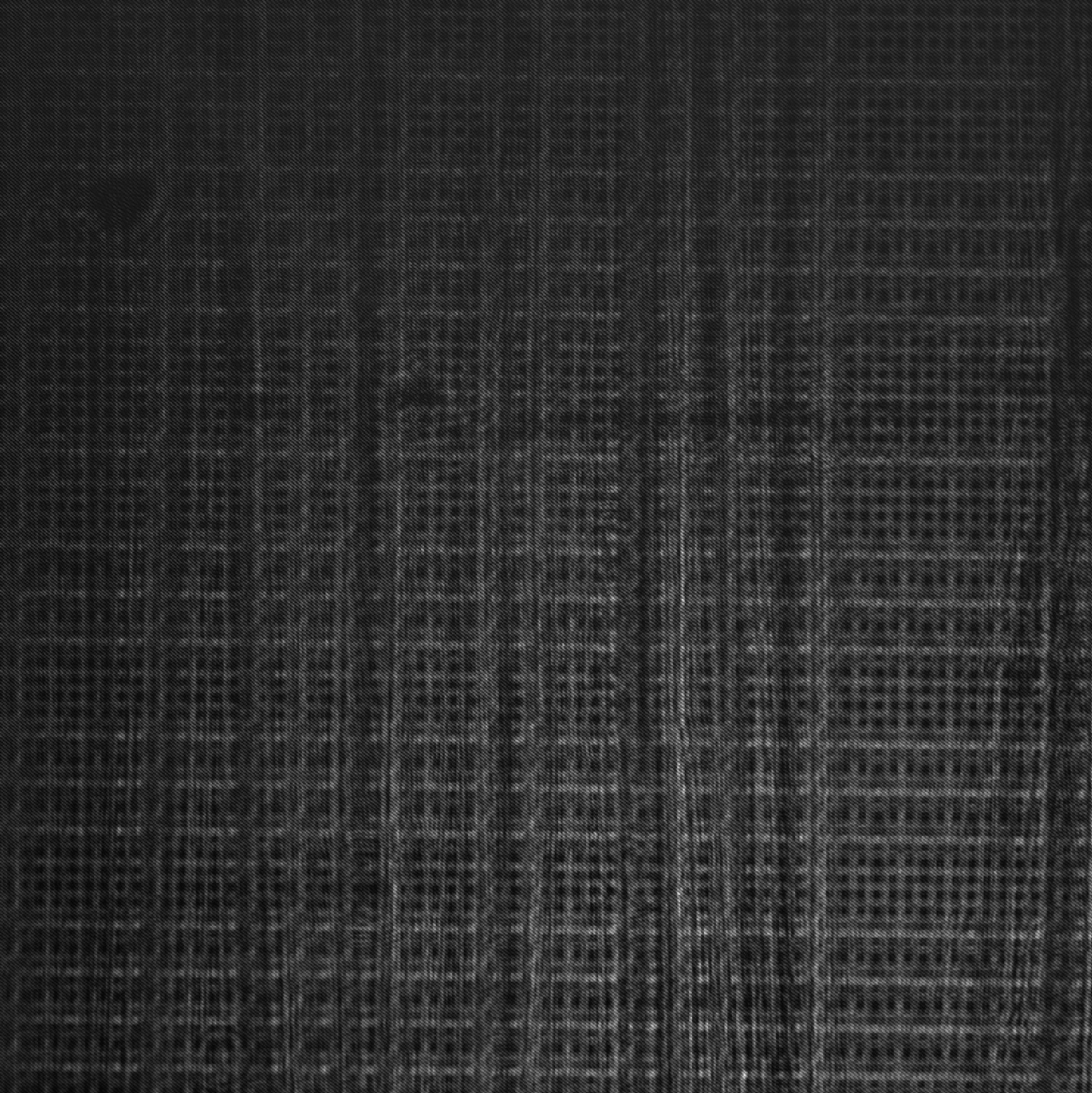}
    \caption{Original hologram (1024x1024)}
    \label{fig:holo_orig}
\end{figure}

The objective is to localize objects at the micrometre scale. We use holograms of a target to retrieve the 6 degrees of freedom (6 DOF) of a structured target.  The target is generated in such a way that it encodes the position (X and Y) using a binary code. The object image needs to be reconstructed first. For that purpose, engineers rely on the known methods which  imply a series of Fourier and inverse  transformations. These transformations consume most of the time and do not allow one to get an idea of the re-constructed object (6 DOF) in a real-time  approach. In this paper, a method is proposed to extract the reconstruction distance Z (which is the distance along the optical propagation) only using deep learning and classification models. As input, produced holograms on a distance Z over $10 \mu m$ are considered. Each hologram (see Figure \ref{fig:holo_orig}) is recorded using the same target. In this paper, this problem is approached using classifier models (ViT \& CNN). For that purpose, the set of holograms has been split in 10 classes. Each class  contains a total of 360 holograms. The digital holography microscope used produces holograms with a resolution equal to 1024x1024.

\begin{figure}[!b]
\subfloat[No Sobel filter (original image) - NFO]{\includegraphics[width=0.5\columnwidth]{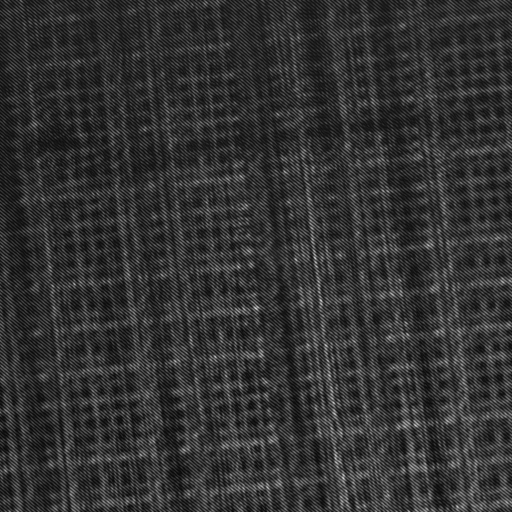}}
\hspace*{\fill}
\subfloat[Sobel filter (original image) - SFO]{\includegraphics[width=0.5\columnwidth]{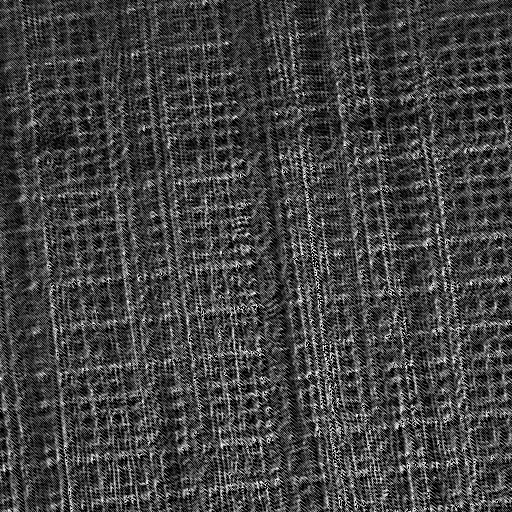}}
\vspace*{\fill}
\subfloat[No Sobel filter (negative image) - NSN]{ \includegraphics[width=0.5\columnwidth]{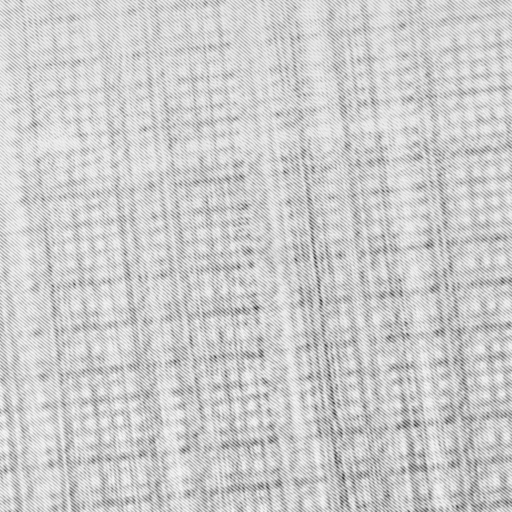}}
\hspace*{\fill}
\subfloat[Sobel filter (negative image) - SFN]{ \includegraphics[width=0.5\columnwidth]{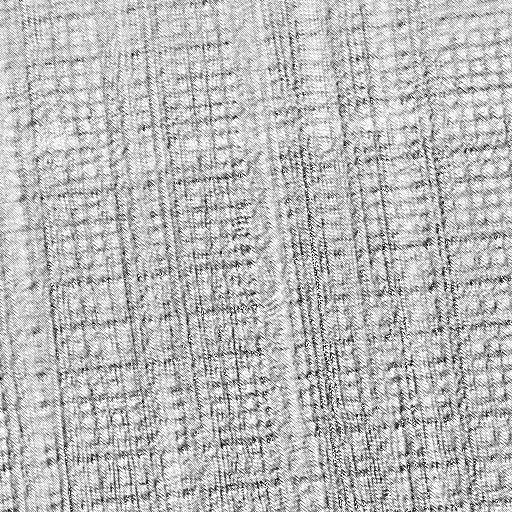}}
\caption{Four different datasets of 512x512 images(holograms) used for training}
\label{fig:compare_holograms}
\end{figure}

A series of deep-learning architectures (CNN \& ViT) has been compared using the following crop approaches to create four datasets of input images (holograms) either taking the original hologram or its negative version (Figure~\ref{fig:compare_holograms}):
\begin{itemize}
    \item A region of interest (ROI) is cropped at the center of the hologram (512x512).
    \item A Sobel filter is applied on the hologram image and a ROI is cropped at the center of the hologram (512x512). 
\end{itemize}

\section{RESULTS}
\begin{figure}[!b]
    \includegraphics[width=1\columnwidth]{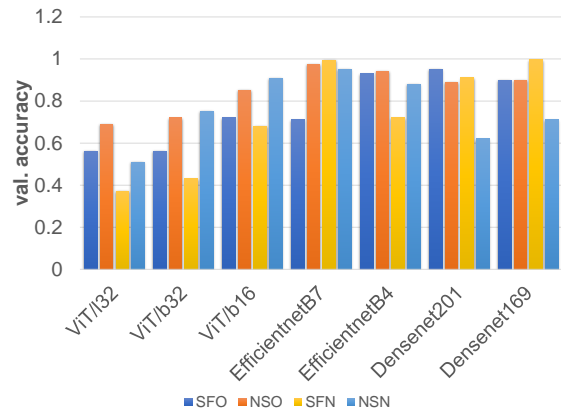}
    \caption{valuation accuracy (val. accuracy) per model architecture}
    \label{fig:training_metrics}
\end{figure}

\begin{figure}[t]
    \subfloat[Test results for the best models (EfficientnetB7, Densenet169 or ViT\_B\_16). A ROI at the center of the original hologram has been considered: in case of a CNN, a Sobel filter is applied on the negative version of the hologram. For ViT\_B16, negative images have been used without any Sobel filter applied.]{\includegraphics[width=1\columnwidth]{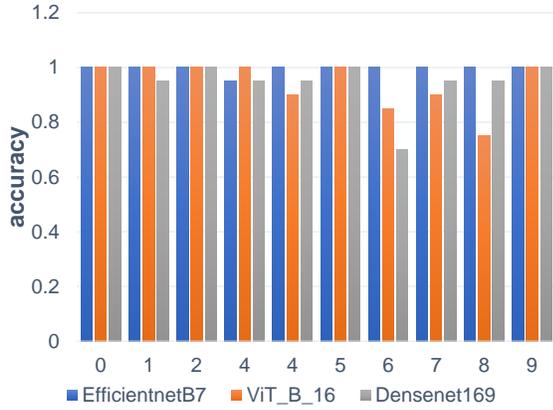}}
  \vspace*{\fill}
    \subfloat[Same test as the  figure (a), but taking ROI in the bottom left corner of the hologram instead at the center.]{\includegraphics[width=1\columnwidth]{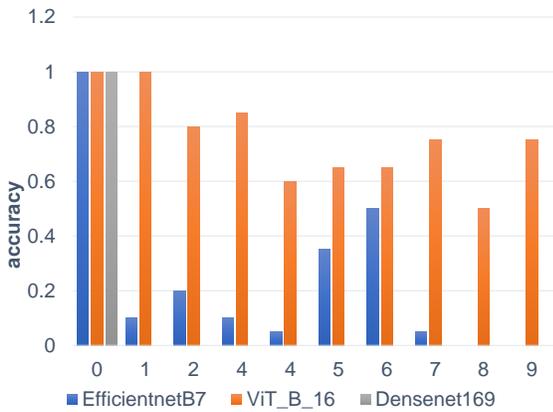}}
    \caption{Test results for the best models: Densenet169, EfficientnetB7 and ViT\_B\_16 over the 10 classes (0 to 9).}
    \label{fig:TestResults}
\end{figure}

All trainings have been executed on a NVidia GPU V100 series. A simple split rule on the input dataset has been applied: 80\% for training and 20\% for validation, basically 3,400 images in total. 200 random images have been extracted from the full dataset of holograms to build our test dataset randomly. Each training has been executed with a max of 200 epochs. An early stop on the valuation loss with a patience of 20 epochs is considered during training. Below the proposed approaches are compared either by applying a Sobel filter on the hologram passed to the neural network or without any filter applied. The results show for each configuration the value of the valuation loss (val. loss) and valuation accuracy (val. accuracy, Figure \ref{fig:training_metrics}) after 200 epochs (Tables \ref{tab:black_results} \& \ref{tab:white_results}). The best models are taken from the comparison and further analyzed to give results during inference on a test dataset (Figure \ref{fig:TestResults}). All the models have been trained with the following frameworks: Tensorflow and Keras. The Adam optimizer has been used with a learning rate of $10^{-4}$. For the CNN, a Global Average Pooling layer has been added before the prediction layer (full connected layer). 

List of architectures compared:
\begin{itemize}
    \item Densenet201 and Densenet169 (CNN)
    \item Efficientnet  B4 and B7 (CNN)
    \item ViT (Vision Transformer) B\_16, B\_32 and L\_32
\end{itemize}
\newpage
4 scenarios are considered for each model (Figure \ref{fig:compare_holograms}). The results are shown in Table \ref{tab:black_results} and Table \ref{tab:white_results}: 

\begin{itemize}
    \item SFO: Sobel filter applied on the original hologram. 
    \item NSO: No sobel filter applied on the original hologram.
    \item SFN: Sobel filter applied on the negative hologram.
    \item NSN: No sobel filter applied on the negative hologram.
\end{itemize}
\begin{table}[t]
\begin{tabular}{lllll}
               & \multicolumn{2}{c}{\textbf{SFO}} & \multicolumn{2}{c}{\textbf{NSO}} \\
               & val. loss    & val. accuracy   & val. loss    & val. accuracy     \\
ViT\_L32        & 0.89         & 0.56       & 0.66          & 0.69         \\
ViT\_B32        & 1.03         & 0.56       & 0.61          & 0.72         \\
ViT\_B16        & 0.88         & 0.72       & 0.36          & 0.85         \\
EfficientnetB7 & 0.69         & 0.71       & 0.08          & 0.9735       \\
EfficientnetB4 & 0.19         & 0.93       & 0.16          & 0.94         \\
Densenet201    & 0.15         & 0.95       & 0.31          & 0.89         \\
Densenet169    & 0.27         & 0.9        & 0.34          & 0.9         
\end{tabular}
\centering
\caption{Test results considering the original images with (SFO) or without Sobel filter (NSO) applied}
\label{tab:black_results}
\end{table}

\begin{table}[t]
\begin{tabular}{lllll}
               & \multicolumn{2}{c}{\textbf{SFN}} & \multicolumn{2}{c}{\textbf{NSN}} \\
               & val. loss        & val. accuracy      & val. loss        & val. accuracy       \\
ViT\_L32        & 1.38             & 0.37           & 1.11             & 0.51            \\
ViT\_B32        & 1.25             & 0.43           & 0.57             & 0.75            \\
ViT\_B16        & 0.67             & 0.68           & 0.22             & \textbf{0.9074}          \\
EfficientnetB7 & 0.03             & \textbf{0.9941}         & 0.13             & 0.95            \\
EfficientnetB4 & 0.59             & 0.72           & 0.33             & 0.88            \\
Densenet201    & 0.24             & 0.91           & 0.75             & 0.62            \\
Densenet169    & 0.05             & \textbf{0.9955}         & 0.61             & 0.71           
\end{tabular}
\centering
\caption{Test results considering negative images with (SFN) or without Sobel filter (NSN) applied. Bold values correspond to best results.}
\label{tab:white_results}
\end{table}

\begin{figure}[htbp]
    \centering
    \includegraphics[width=0.9\columnwidth]{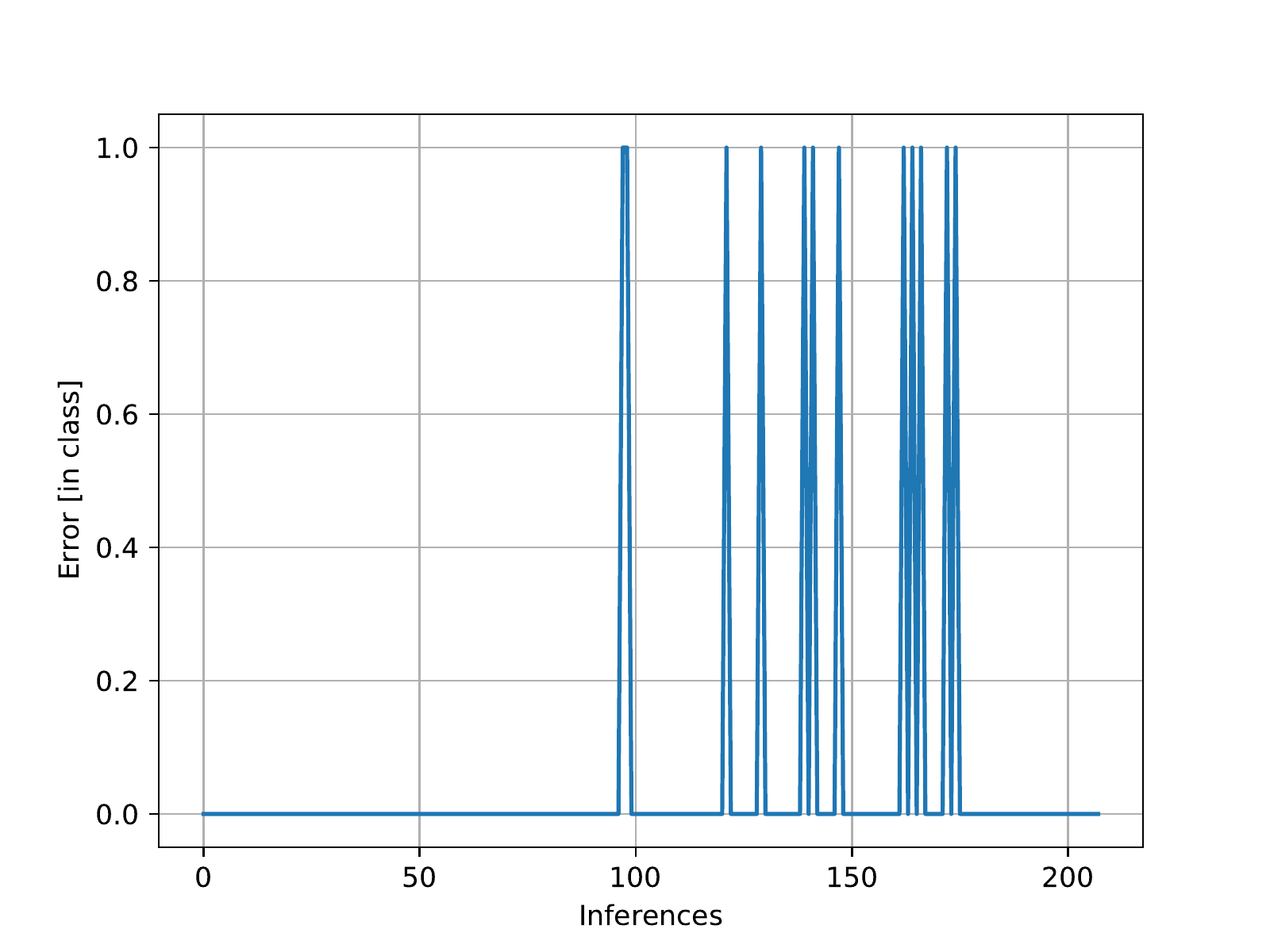}
    \caption{ViT\_B16 inference results considering images cropped at the center of the input hologram}
    \label{fig:vit_error}
\end{figure}

\begin{figure}[htbp]
    \centering
    \includegraphics[width=0.9\columnwidth]{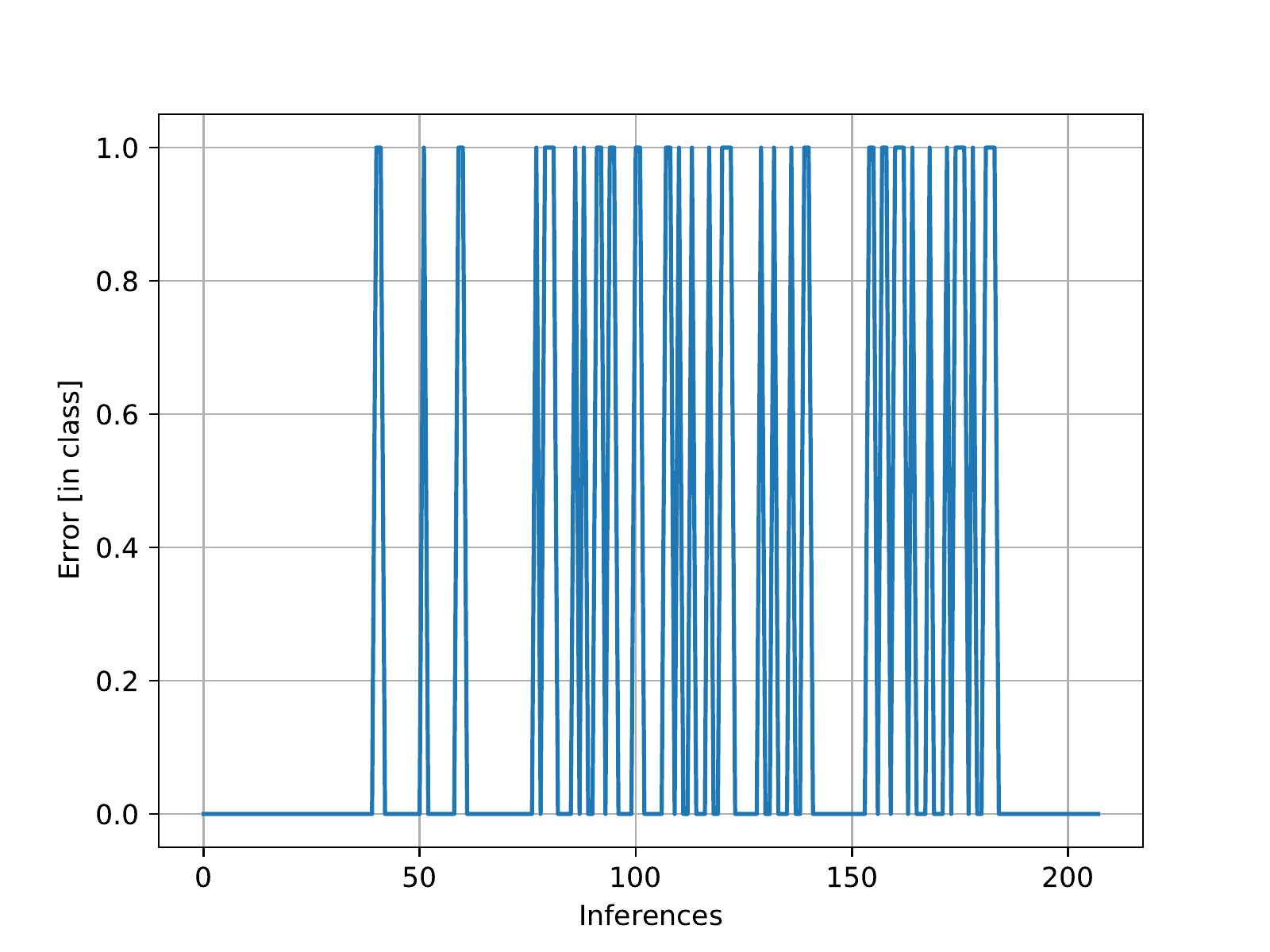}
    \caption{ViT\_B16 inference results considering images cropped in the bottom left corner of the input hologram}
    \label{fig:vit_error_off_region}
\end{figure}

\begin{figure}[htbp]
    \centering
    \includegraphics[width=0.9\columnwidth]{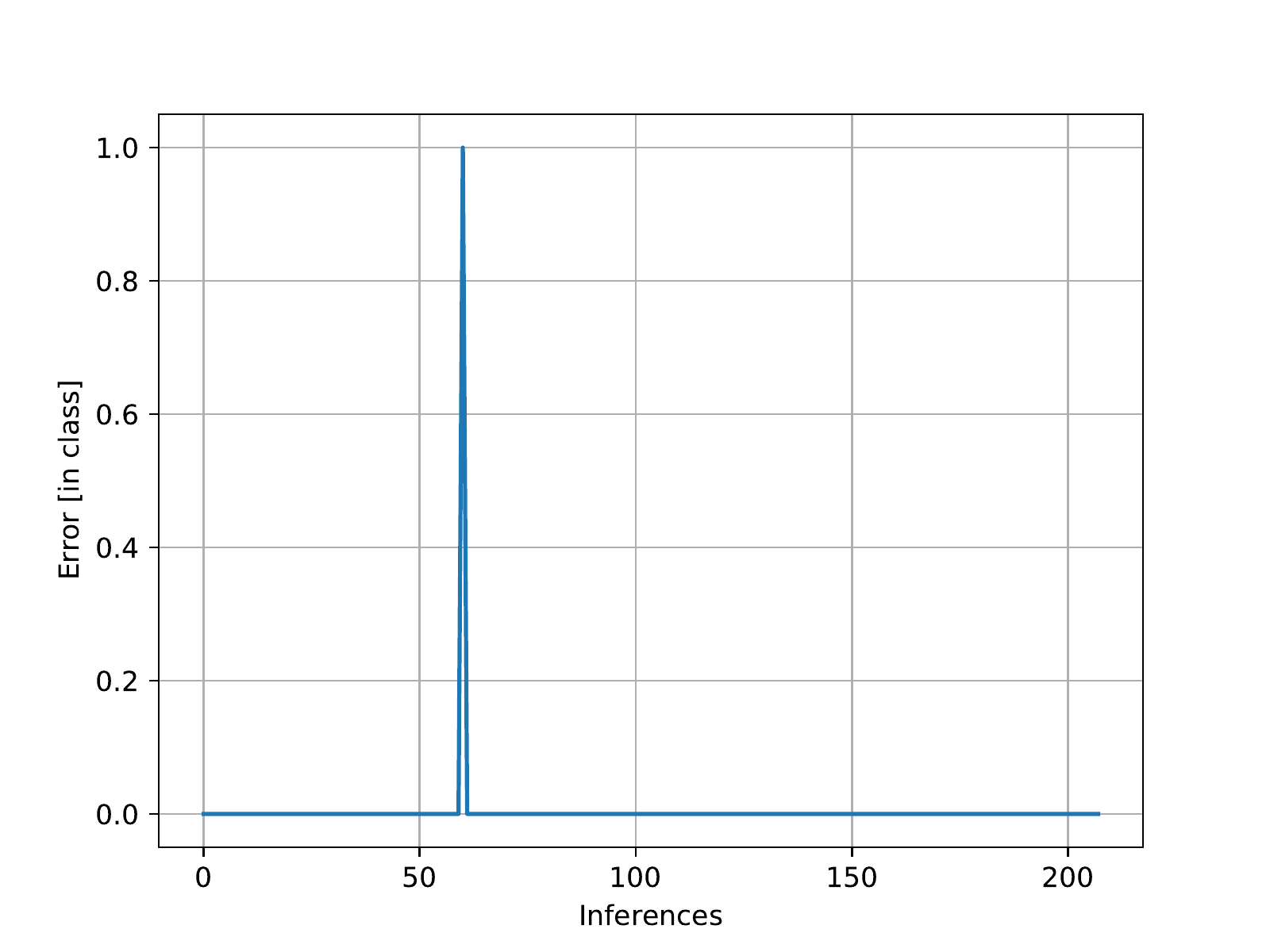}
    \caption{EfficientnetB7 inference results considering images cropped at the center of the input hologram}
    \label{fig:cnn_error1}
\end{figure}

\begin{figure}[htbp]
    \centering
    \includegraphics[width=0.9\columnwidth]{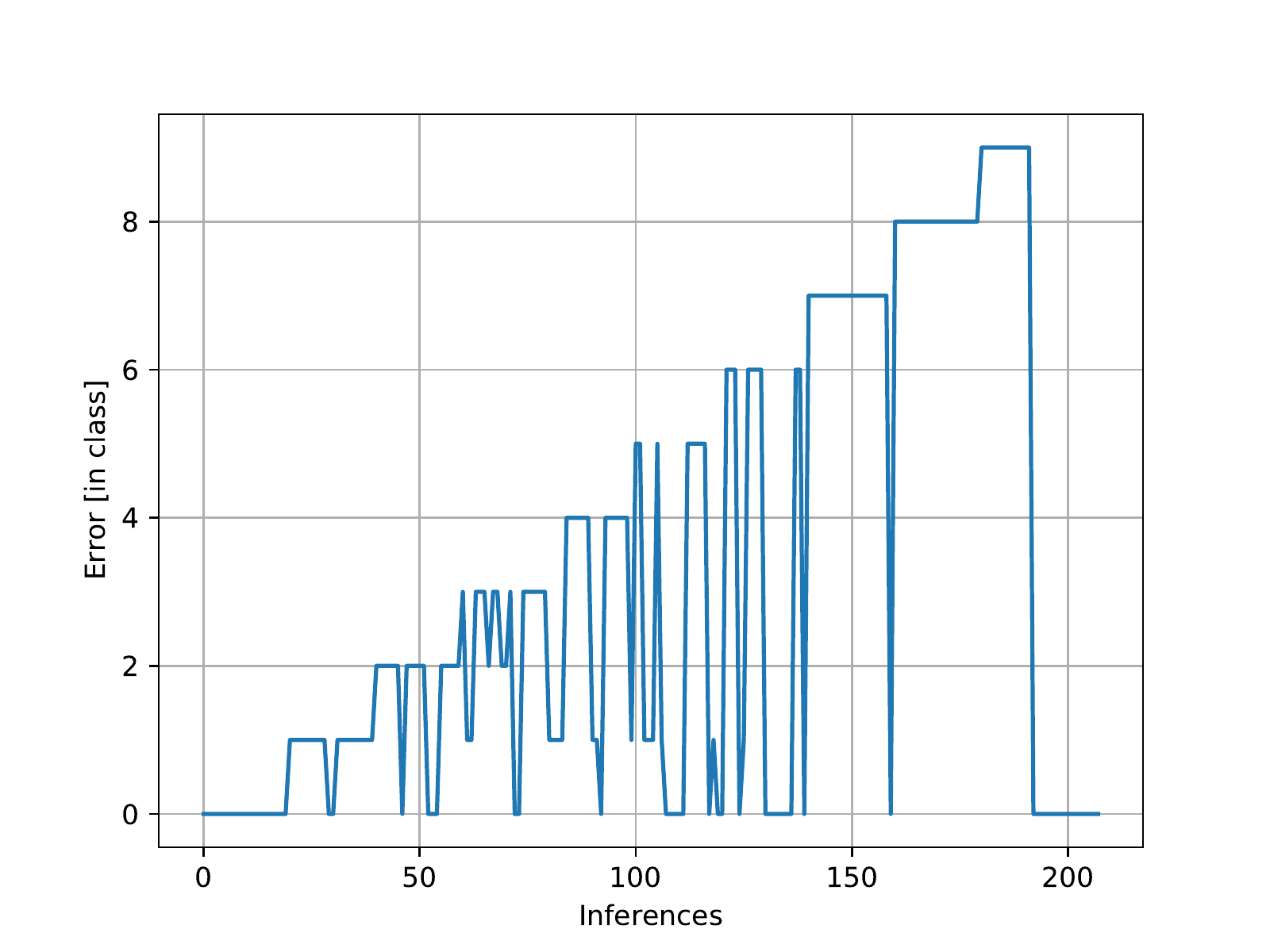}
    \caption{EfficientnetB7 inference results considering images cropped in the bottom left corner of the input hologram}
    \label{fig:cnn_error4}
\end{figure}

\begin{figure}[htbp]
    \centering
    \includegraphics[width=0.9\columnwidth]{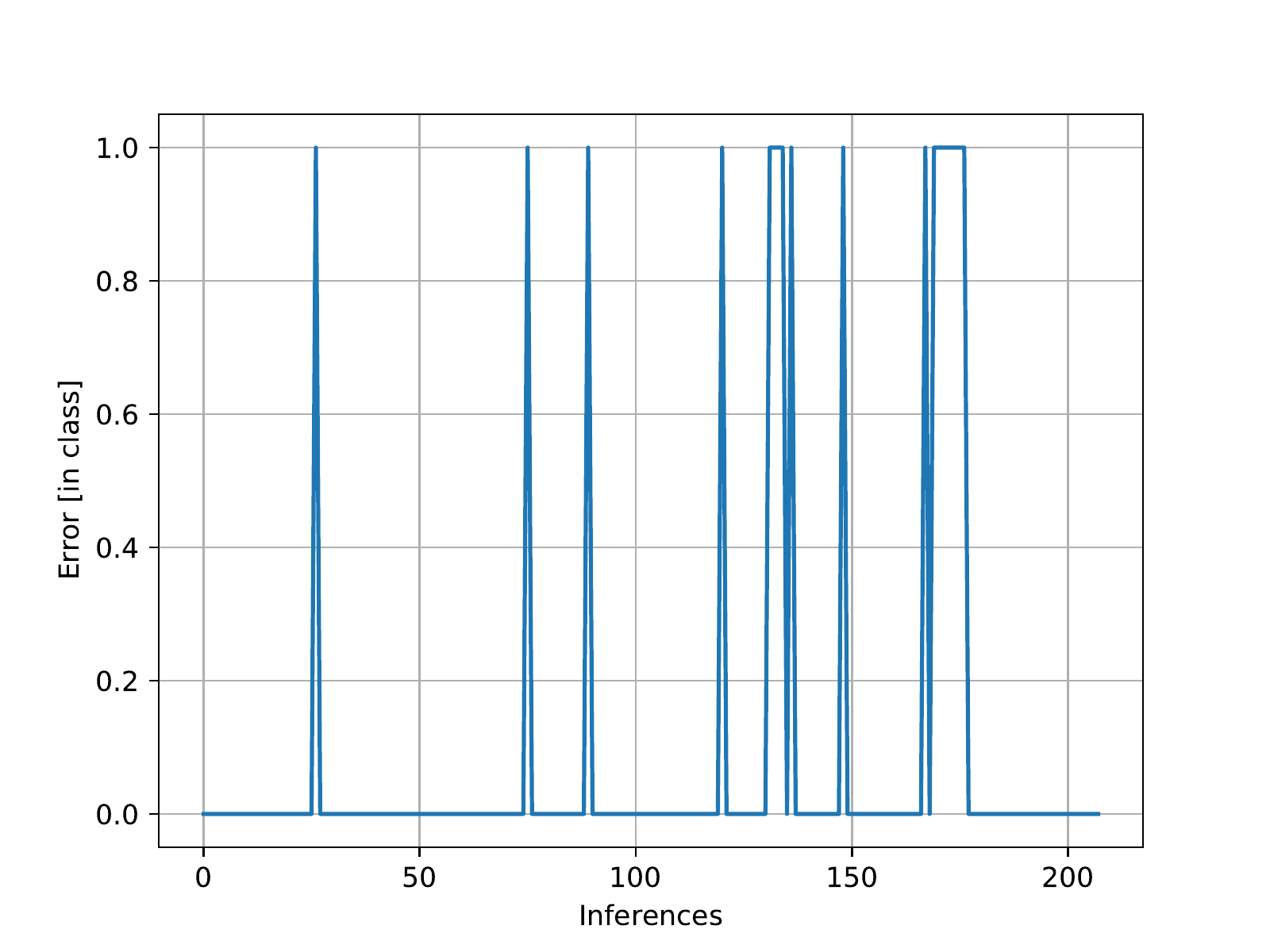}
    \caption{Densenet169 inference results considering images cropped at the center of the input hologram}
   \label{fig:cnn_error2}
\end{figure}

\begin{figure}[htbp]
    \centering
    \includegraphics[width=0.9\columnwidth]{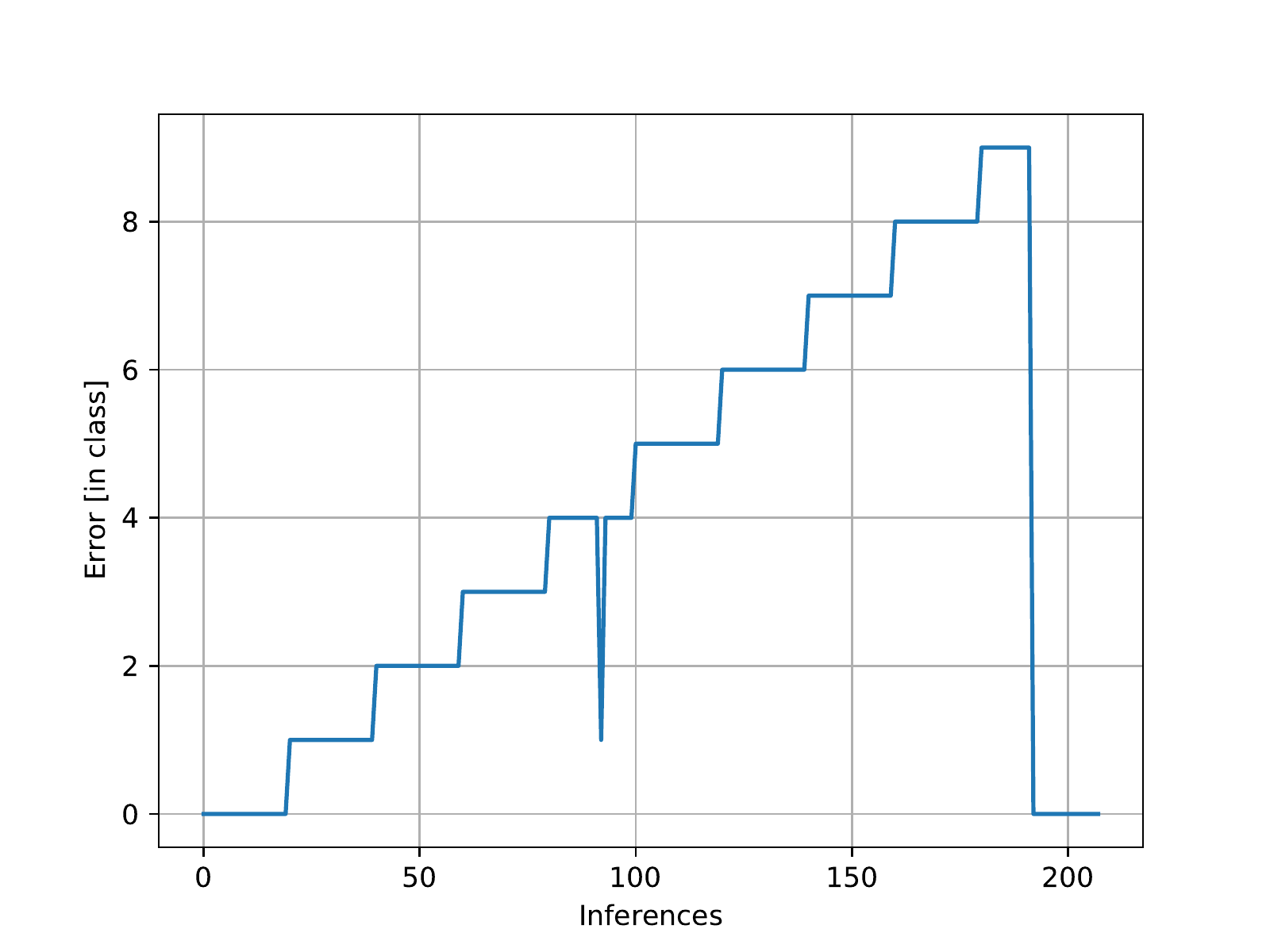}
    \caption{Densenet169 inference results considering images cropped in the bottom left corner of the input hologram}
    \label{fig:cnn_error3}
\end{figure}

Below, the performance on the test dataset for the best models trained with Densenet169, EfficientnetB7 and ViT\_B\_16 are analyzed (Figure \ref{fig:TestResults}). The inference is executed using a test dataset with a total of 200 images, 20 images per class. Each image per class is selected randomly to minimize the chance effect during inference. 
The error (distance error in $\mu$m) during inference is evaluated taking ROI pointing the same region as the one taken during training or the bottom left corner (Figures \ref{fig:vit_error}, \ref{fig:vit_error_off_region}, \ref{fig:cnn_error1}, \ref{fig:cnn_error4}, \ref{fig:cnn_error2} and \ref{fig:cnn_error3}).
This error, independently of the chosen architecture either Densenet169, EfficientnetB7 or ViT\_B16, is bounded by $1-3 \mu$m.
In case of ViT\_B16, a slightly higher number of errors is visible due to the lowest performance of the model. Even if the model is pointing a different zone than the one used during training, ViT\_B16 performs well and the max absolute error in number of classes is of 1 class ($1-3 \mu$m).
\section{ViT vs CNN}
\begin{figure}[htbp]
    \subfloat[Grad-Cam EfficientnetB7 (sample 1) ]{\includegraphics[width=0.4\columnwidth]{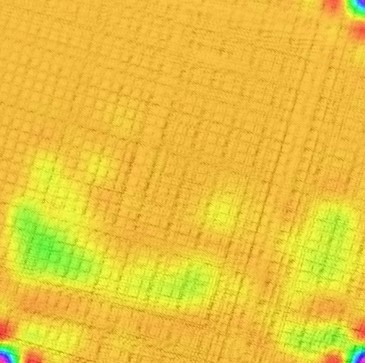}}
    \vspace*{\fill}
    \subfloat[Grad-Cam EfficientnetB7 (sample 2)]{\includegraphics[width=0.4\columnwidth]{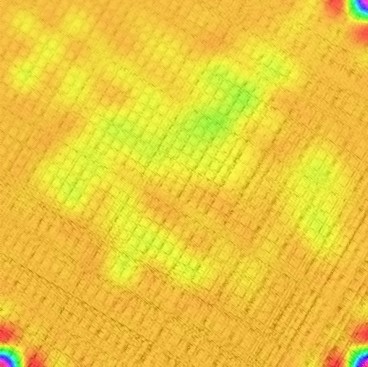}}
    \vspace*{\fill}
    \subfloat[Grad-Cam Densenet169 (sample 1)]{\includegraphics[width=0.4\columnwidth]{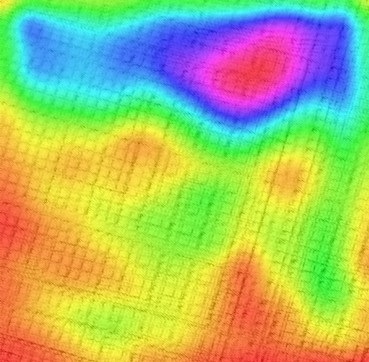}}
    \vspace*{\fill}
    \subfloat[Grad-Cam Densenet169 (sample 2)]{\includegraphics[width=0.4\columnwidth]{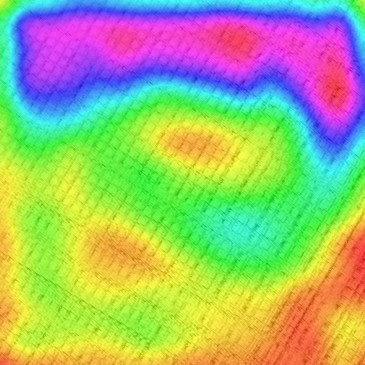}}
    \vspace*{\fill}
    \subfloat[Attention map ViT\_B\_16 ]{\includegraphics[width=0.4\columnwidth]{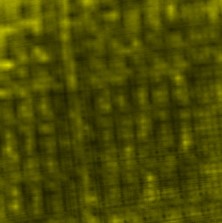}}
    \caption{Grad-Cam (CNN) and attention map (ViT) of the best architecture (EfficientNetB7, Densenet169 \& ViT\_B\_16)}
    \label{fig:model_focus}
\end{figure}

According to our results and considering the same ROI as the one used during training, the best architecture for our purpose seems to be EfficientnetB7. Such a model achieves a better performance compared to a model trained using ViT. A ViT network should perform as well as a residual neural network such as Densenet or Efficientnet \cite{b15}, even reaching a higher performance. The reason must be linked to how these neural networks see the holograms (images passed to the neural network) during training. Different figures are proposed as shown on  Figure \ref{fig:model_focus}, Grad-Cam's (as described in \cite{b17}) for CNN and the attention map for ViT.

Grad-cam is giving an idea of where a CNN network is looking during training. The figure shows where the neural network focuses after the last convolution layer (block7d\_project\_conv for EfficientnetB7 and conv5\_block32\_2\_conv for Densenet169) before batch normalization and prediction layer (dense layer). The network seems to focus its attention on specific location inside the hologram. In case of Densenet169, the CNN strongly focuses on these specific locations inside the hologram. For ViT models, the attention map (Figure \ref{fig:model_focus}) is generated. This demonstrates that the model tries to look globally inside the hologram which significantly influences the performance of the model during the training and the inference. It seems that ViT is slightly less accurate during training (val.loss and val. accuracy) than CNN (Densenet169 or EfficientnetB7). Densenet169 focuses more on specific asperities which explains the obtained results (Figure \ref{fig:training_metrics}), the valuation accuracy reaches an higher value compared to EfficientnetB7 or ViT\_B16. It seems that the CNN architecture learns the shape of the target more than the physical phenomena. During inference and taking the same ROI for the test dataset, we can see that EfficientnetB7 performs slightly better than Densenet169. This goes in the direction of our assumption as EfficientnetB7 tries to consider the hologram more globally during training. Taking a different ROI, ViT\_B16 achieves a better performance. This is totally explainable as the attention map shows, ViT\_B16 is learning from the whole hologram and not specific locations (this is due to the self-attention \cite{b16}). ViT models seem to focus their attention on the phenomena more than the target. The general performance of the model is slightly impacted, but the max absolute error stays in an acceptable range bounded by $1-3\mu$m.
\section{Conclusions}
Our experiments showed that the reconstruction distance can be found for digital holography using deep learning techniques, especially classification models. The $1-3 \mu$m scale has been reached for a dataset of holograms on a scale of $10 \mu$m. This solution allows to surpass by a factor 2 to 3 the optical resolution of the microscope (using a 10x microscope objective) knowing that the optical resolution is defined as: $r = \lambda /(NA)^2 = 6.6 \mu m$ (with NA = 0.3: numerical aperture and $\lambda=0.6 \mu m$). The max error of one class stays below or equal to $1-3\mu m$ when the model fails to recognize the right distance. ViT models are more robust when pointing arbitrary ROI than CNN networks.  
In future work, our focus will be on regression models. A regression might help to reach a higher accuracy as the problem of auto-focusing is not a discrete, but a continuous function. 

\section*{Acknowledgment}

This work was supported by ROBOTEX (ANR-10-EQPX-44-01), by Cross-disciplinary Research (EIPHI) Graduate School (contract ANR-17-EURE-0002), the French Investissements d’Avenir program, the I-SITE Bourgogne Franche-Comté (BFC) project (contract ANR-15-IDEX-03) and Région Bourgogne Franche-Comté (ControlNet project).
This work was performed using HPC resources from GENCI-IDRIS (Grant 20XX-AD011012913).



\begin{thebibliography}{00}
\bibitem{b1}  R. Couturier, M. Salomon, E. A. Zeid, and C. A. Jaoudé. Using deep learning for object distance prediction in digital holography. In 2021 International Conference on Computer, Control and Robotics (ICCCR), pages 231–235, 2021.
\bibitem{b11} M. K. Kim. Principles and techniques of digital holographic microscopy. SPIE Reviews, 1(1):1 – 51,2010.
\bibitem{b2}  I. Acharya and D. Upadhyay. Comparative study of digital holography reconstruction methods. Procedia Computer Science, 58:649–658, 12 2015.
\bibitem{b3}  A. Hong, B. Zeydan, S. Charreyron, O. Ergeneman,S. Pané, M. F. Toy, A. J. Petruska, and B. J. Nelson.Real-time holographic tracking and control of microrobots. IEEE Robotics and Automation Letters,2(1):143–148, 2017.
\bibitem{b4}  Z. Ren, Z. Xu, and E. Y. Lam. Autofocusing in digital holography using deep learning. In T. G. Brown, C. J.Cogswell, and T. Wilson, editors, Three-Dimensional and Multidimensional Microscopy: Image Acquisition and Processing XXV, volume 10499, pages 157 – 164. International Society for Optics and Photonics, SPIE,2018.
\bibitem{b5}  Z. Ren, Z. Xu, and E. Lam. Learning-based non parametric autofocusing for digital holography. Optica, 5:337, 04 2018.
\bibitem{b6}  H. A. Ilhan, M. Dogar, and M. \"Ozcan. Digital holographic microscopy and focusing methods based on image sharpness. Journal of Microscopy, 255(3):138–149, 2014.
\bibitem{b7} Z. Ren, Z. Xu, and E. Lam. Learning-based non parametric autofocusing for digital holography. Optica, 5:337, 04 2018.
\bibitem{b8}  T. Shimobaba, T. Kakue, and T. Ito. Convolutional neural network-based regression for depth prediction in digital holography. In 2018 IEEE 27th International Symposium on Industrial Electronics (ISIE), pages1323–1326, 2018.
\bibitem{b9}  T. Pitkaaho, A. Manninen, and T. J. Naughton. Performance of autofocus capability of deep convolutional neural networks in digital holographic microscopy. In Digital Holography and Three-Dimensional Imaging, page W2A.5. Optical Society of America, 2017.
\bibitem{b10}  T. Pitkaaho, A. Manninen, and T. J. Naughton. Focus classification in digital holographic microscopy using deep convolutional neural networks. In E. Beaurepaire, F. S. Pavone, and P. T. C. So, editors,Advances in Microscopic Imaging, volume 10414, pages 89 – 91. International Society for Optics and Photonics, SPIE, 2017.
\bibitem{b12} M. Gu, Shanqing; Pednekar and R. Slater. Improve image classification using data augmentation and neural networks. SMU Data Science Review, 2(2),2019.
\bibitem{b13}  G. Huang, Z. Liu, L. Van Der Maaten, and K. Q. Weinberger. Densely connected convolutional networks. In2017 IEEE Conference on Computer Vision and Pattern Recognition (CVPR), pages2261–2269, 2017.
\bibitem{b14}  R. Kumar. Adding binary search connections to improve densenet performance. 5th International Conference on Next Generation Computing Technologies (NGCT-2019), 2020.
\bibitem{b15} A. Dosovitskiy, L. Beyer, A. Kolesnikov, D. Weissenborn, X. Zhai, T. Unterthiner, M. Dehghani, M. Minderer, G. Heigold, S. Gelly,J. Uszkoreit, and N. Houlsby. An image is worth 16x16 words:  Transformers for image recognition at scale, 2020. arXiv:2010.11929.
\bibitem{b17} R. R. Selvaraju, M. Cogswell, A. Das, R. Vedantam,D. Parikh, and D. Batra. Grad-cam:  Visual explanations from deep networks via gradient-based localization. International Journal of Computer Vision, 128(2):336–359, Oct 2019.
\bibitem{b16}  A. Vaswani, N. Shazeer, N. Parmar, J. Uszkoreit,L. Jones, A. N. Gomez, L. Kaiser, and I. Polosukhin. Attention is all you need, 2017. arXiv:1706.03762.
\end{thebibliography}
\end{document}